\documentstyle[a4wide,twocolumn]{article}

\title{How much is enough?: \\
       Data requirements for statistical NLP}

\author{Mark Lauer \\
Microsoft Institute \\
65 Epping Road, North Ryde NSW 2113 \\
Australia \\
{\tt t-markl@microsoft.com}
}

\date{}

\begin{document}

\maketitle

\begin{abstract}
Feeding training data to statistical representations of language
has become a popular past-time for computational
linguists, but our understanding of what constitutes
a sufficient volume of data remains shadowy.
For example, Brown {\it et al.} (1992) used over 500 million
words of text to train their language model.
Is this enough?  Could devouring even more data
further improve the accuracy of the model learnt?
In this paper I explore a number of issues in the analysis of
data requirements for statistical NLP systems.\footnote{This
paper has been published in the Proceedings of the Second
Conference of the Pacific Association for Computational
Linguistics, Brisbane, Australia, 1995.}
A framework for viewing such systems is proposed and a sample
of existing works are compared within this framework.
Finally, the first steps toward a theory of
data requirements are made by establishing
an upper bound on the expected error rate of a class of
statistical language learners as a function of the volume
of training data.
\end{abstract}

\section{Introduction}

Statistical approaches to natural language processing
are becoming increasingly popular, being applied
to a wide variety of tasks.  For example,
Weischedel {\it et al.} (1993) explores part-of-speech
tagging, parsing and acquisition of lexical frames.
Nonetheless, all these tasks share some important
characteristics, not the least of which is the requirement
for a sizable corpus of training data.  One
question which has largely been ignored is how much
data is enough?  For example, given a limited body
of training data, it is essential to know which
statistical NLP methods are likely to be accurate before
pursuing any one.  Also, given a particular
method, when will acquiring further training data cease
to improve the system accuracy?  Currently, the
field is conspicuously lacking a general theory of
data requirements for statistical NLP.

In this paper, I present the first steps towards the
development of such a theory.  I begin by formulating
a framework for statistical NLP systems
designed to capture some of the elements crucial
to data requirements analysis.  I will
then review a sample of existing approaches, showing
how they fit into the framework, and where they vary from it.
Even though several of these introduce complexities which
are not captured by the framework, reasoning in the framework
still supports some important insights into these systems.
Finally, I present some preliminary work on establishing
a closed form upper bound on data requirements for a class
of statistical NLP systems. This latter work owes much to Mark
Johnson, Brown University, who is responsible for several key
mathematical ideas in Section \ref{theory}.

Statistical NLP systems are designed to make choices;
hopefully in an informed manner.  To do this they use
indicators, upon which their choices are conditioned.
The purpose of computing statistics is to inductively establish
the relationship between the indicators and the choices to be made.
Consider for example a next word predictor which attempts to predict
the next word on the basis of the preceding word.  To do this it must
have an understanding of the relationship  between the indicator (the
preceding word) and the choice (the next word).  It is possible
to acquire this understanding by computing statistics over a large
corpus, a process called training.  Once trained, the system may then
be applied to a new text and its accuracy evaluated.  This paper
is concerned with the dependence of a system's accuracy
on the size of the training corpus.  In the following section, the
notions of indicators, choices and training data will
be made more formal.

\section{Statistical Processors}
\subsection{A Framework}

A statistical NLP system deals with a certain linguistic
universe.  Formally, there is a set of linguistic events
$\Omega$ from which every training example and every
test instance will be drawn.  In the next word predictor,
this need only be the set of all pairs of words which may
be adjacent in text.

Let $V$ be a finite set of values that we would like
to assign to a given linguistic input.  This defines
the range of possible answers that the analyser can
make.  In the predictor, this is the set of words plus
an end of sentence symbol.
Let $J:\Omega \rightarrow V$ be the random
variable describing the distribution of values that
linguistic events take on.  We also require a set of
indicators, $B$, to use in selecting a value for a
given linguistic event.  I will refer to each element of
$B$ as a {\em bin}. In the predictor, the set of bins is
the set of words plus a start of sentence symbol.
Let $I:\Omega \rightarrow B$ be
the random variable describing the distribution of bins
into which linguistic events fall.

The task of the analyser is to choose which value
is most likely given only the
indicator.  Therefore, it is a function $A:B \rightarrow
V$.  The task of the learning algorithm is to acquire
this function by computing statistics on the training set.

Putting these components together, we can define a
{\em Statistical Processor}, \cal{S} as a
tuple $ \langle \Omega , B , V , A \rangle $, where:
\begin{itemize}
\item $\Omega$ is the set of all possible linguistic
events
\item $B$ and $V$ are finite sets, the bins and values
respectively
\item $A$ is the trained analysis function
\end{itemize}

Amongst all such statistical processors, there is a
special class in which we are interested.  Define a
{\em probabilistic analyser} to be a statistical processor
which computes a function $p:B \times V
\rightarrow [0,1]$ such that $\sum_{v \in V} p(b,v)
= 1$ for all $b \in B$ and then computes $A$ as:
\begin{equation}
A(b) = \mbox{argmax}_{\,v \in V} p(b,v)
\end{equation}
The problem of acquiring $A$ is thus transformed into
one of estimating the function $p$ using the
training corpus. Generally, $p(b,v)$ is viewed as an estimate
of the probability $\Pr(J=v | I=b)$.

\subsection{Training Data}

Formally, a training corpus, $c$,
of $m$ {\em instances}, is an element from
$(B \times V)^{m}$ where each pair $(b,v)$ is sampled
according to the random variables $I$ and $J$ from
$\Omega$.  For probabilistic analysers, there are a variety of
methods by which an appropriate function $p$ can be
estimated from a corpus; one simple example being the
Maximum Likelihood Estimate.  Regardless of
the learning algorithm used, each possible training
corpus, $c$, results in the acquisition of some function, $P_{c}$.
Our aim is to explore the dependence
of the expected accuracy of $P_{c}$ on the magnitude
of $m$.

Surprisingly, it is not always obvious how many training
instances have been used to train a statistical
method.  It is not generally sufficient
to report the size of the corpus in words.  A system
which collects word associations using a window of
cooccurrence 10 words wide will find 819 instances
in a 100 word corpus, while one collecting the objects
of the preposition {\em on} from the same corpus,
would most likely find only a few instances.
Therefore, before any conclusions can be drawn about data
requirements, the training corpus must be measured in terms of
instances.

Each of these instance falls into a particular bin
by virtue of its associated indicator.  In choosing the indicators,
we have implicitly defined equivalence classes for instances.  The
statistical processor will treat every instance in a bin identically.
Further, once the bins are chosen, the greater the number of
training instances that fall into a bin, the greater our
confidence in the statistical inference made by the processor
for test cases in that bin.  For instance, the next word predictor
is more likely to be correct when the preceding word is common
than when it is a rare word.

It is not always obvious how many bins a
given statistical method employs.  Often multiple
indicators are used.
For instance, a trigram tagger uses the tags of the two
preceding words and the current word to choose a new tag.
In this case, $B = T \times T \times W$ where $T$ is the tagset
and $W$ is the vocabulary.

This example demonstrates an important point.  By choosing to take
into account the tags of two preceding words, the trigram
tagger requires $|T|$ times as many bins as a bigram
tagger (where $B = T \times W$).  With more bins, the
trigram tagger is sensitive to a broader range of context and
thus can in principle achieve a greater accuracy.  However,
because there are more bins, there are fewer training instances
in each bin.  Thus, statistical estimation will be less accurate.
In practice, high accuracy requires at least a few training
instances per bin.  Thus increasing the number of indicators
may actually decrease the overall accuracy.

For probabilistic analysers it is useful to define the number
of {\em slots}, $L$, to be $|B|(|V|-1)$, which is the number of
independent parameters needed to define the function $p$.

\subsection{Error Rates and Optimality}

For any non-trivial general statistical processor the
indicators used cannot perfectly represent the entire
linguistic event space.  Thus, in general there exist
values $v_{1}, v_{2} \in V$, for which
both $\Pr(J=v_{1} | I=b) > 0$
and $\Pr(J=v_{2} | I=b) > 0$ for some $b \in B$.
Suppose without loss of generality that
$A(b) = v_{1}$.  The analyser will be in
error with probability at least $\Pr(J=v_{2} , I=b)$.
This is the root of a rather difficult problem in
statistical NLP because no matter how inaccurate a
trained statistical processor is, the inaccuracy may
be due to the imperfect representation of $\Omega$
by $B$.  \footnote{Unless a more accurate statistical
processor based on the same indicators already exists.}

Probabilistic analysers always select just one value for each bin,
the one which maximises $p$.  Let $v_{\mbox{\it mode}}(p, b)
= \mbox{argmax}_{v \in V} p(b,v)$.
This leads to a definition for the expected error rate
of a function $p$, $R(p)$:
\begin{eqnarray} \label{eer_def}
\lefteqn{R(p) =} \\
\lefteqn{\sum_{b \in B} \Pr(I=b) ( \!\!\!\!\!\!\!\!\!\!\!\!
\sum_{v \in V \setminus \{ v_{\mbox{\it mode}}(p, b) \} }
\!\!\!\!\!\!\!\!\!\!\!\! \Pr(J=v | I=b) )} \nonumber \\
& \hspace{6cm} & \nonumber
\end{eqnarray}

This is the probability of the analyser being in error
on a randomly selected element of $\Omega$.  Let
$p_{\mbox{\it opt}}$ be any function which minimises the expected
error rate and $r_{\mbox{\it opt}} = R(p_{\mbox{\it opt}})$.
Given $B$ and $V$, $r_{\mbox{\it opt}}$ is the smallest possible
expected error rate.  Any probabilistic analyser
which achieves an accuracy close to this is
unlikely to benefit from further training data.

Unless large volumes of manually annotated data exist,
measuring the size of $r_{\mbox{\it opt}}$ in any given statistical
processor presents a difficult challenge.  Hindle
and Rooth (1993) have attempted a similar task using human
subjects on the problem of prepositional phrase
attachment.  Subjects were given only the preposition
and the preceding verb and noun and then were
asked to select the attachment.  This was precisely
the task facing their statistical processor.  The subjects
could only perform the attachment correctly in around 86\%
of cases.  If we assume that the subjects incorrectly analysed
the remaining 14\% of cases because these cases depended on
knowledge of the wider context, then any statistical learning
algorithm based only on these indicators cannot do better
than 86\%.  Of course, if there is insufficient training data
the system may do considerably worse.

Assuming that human performance on the task
accurately reflects the value of $r_{\mbox{\it opt}}$
is the only means known at present to estimate the value
of $r_{\mbox{\it opt}}$.
Unfortunately, this approach is expensive to apply
and makes a number of questionable
psychological assumptions.  For example,
it assumes that humans can accurately reproduce parts
of their language analysis behaviour on
command.  It may also suffer when representational
aspects of the analysis task cannot be explained
easily to experimental subjects.  A worthwhile goal
for future research is to establish a statistical
method for estimating or bounding  $r_{\mbox{\it opt}}$ using
language data.

\section{Statistical Learning}
\subsection{Existing Methods}

In this section, I show how a number of existing statistical
NLP works fit into the framework, including a
tagger, a sense disambiguator and three syntactic analysers.
 For each, I consider how the various
elements of the general statistical processor are instantiated.

Weischedel {\it et al.} (1993) uses (among other experiments)
a trigram hidden Markov model to tag text for
part of speech.  The training data is four million
words of the University of Pennsylvania Treebank,
tagged with a set of 47 different tags.  I shall regard
$B$ as consisting of the two previous tags ($T \times T$),
while $V$ is simply the tagset.  The system
also takes into account lexical tag frequencies (that is,
$B = T \times T \times W$).
I will assume however that data sparseness does not affect
the lexical tag frequency estimates.  Since the trigram estimates and
the lexical tag frequencies are combined as independent
factors, ignoring the lexical component does not seem unreasonable.
The situation is further complicated
because probability is maximised over a sequence of
words, rather than for a single word.  The framework needs to be extended
to capture these mechanisms, but for the moment the approximations
I have made may be useful.
Since every word in the corpus (bar the first two) is used for
training, we have $m=4$ million and $L= 47 \times 47
\times 46$.  The accuracy is reported to be
around 97\%, which is approximately the accuracy of
human taggers using the whole context.

Yarowsky (1992) describes a sense disambiguation system
which uses a 100 word window of
cooccurrences.  He uses a mutual information-like measure
which combines the cooccurrence statistics
for all words in each category of Roget's thesaurus.
 The result is a profile of contexts for a category
which can be used to estimate how likely each category
is within a certain context.  Comparing the
different possible categories for the word provides
a broad sense discrimination.  The training corpus is
Grolier's encyclopedia which contains on the order
of 10 million words.  Each of these provides 100
training instances (every other word in the window),
so $m \approx 1$ billion.  Since the evidence from
each word in the context is combined independently,
it is reasonable to regard $B$ as simply the set of distinct
words in Grolier's. Again, further work is needed to make this
approximation unnecessary.  $V$ is the set of Roget's categories
($|V| = 1042$), so assuming the vocabulary is around
100,000, $L \approx 100$ million.
\footnote{Some stemming is performed, so it is the number
of stems in the vocabulary that we want.}
The average accuracy reported is 92\%.

Hindle and Rooth (1993) propose a system to syntactically
disambiguate prepositional phrase
attachments.  Unambiguous examples of attachments are
used to find lexical associations (a likelihood
ratio) between prepositions and the nouns or verbs
they attach to.  They cyclically apply this
technique, adding disambiguated attachments into the
training set, until all the training data (ambiguous
or not) has been used.  This approach can be
approximated by a probabilistic analyser.
Each association value is ascribed
to a pair $(w,p)$ where $w$ is a verb or noun
and $p$ is a preposition.  Thus $B$ is a product of
two indicator spaces: the set of verbs and nouns and
the set of prepositions.  Assuming they used 10,000
nouns and verbs (5,000 of each) and 100
prepositions, $|B| = 1$ million.  The analyser computes
a probability for each of two possible attachments,
nominal and verbal, so $V$ is binary.
The training set consists of 754,000 noun attachments
and 468 thousand verb ones giving $m = 1.22$ million.
\footnote{I have allowed all the training
examples as instances, even though some are
acquired by cyclic refinement.} The accuracy reported
is close to 80\%, while human subjects given the
same indications could achieve 85--88\% accuracy.
If the latter figure reflects the optimal error
rate, it appears there is still room for improvement
by adding training data or changing the statistical
measures.

Lauer (1994) describes a system for syntactically analysing
compound nouns.  Two-word compounds
extracted from Grolier's encyclopedia were used to
measure mutual information between every pair of
thesaurus categories (using Roget's thesaurus) and
the results used to select a bracketing for three-word
compounds.  Since an association value is computed
for every pair of thesaurus categories, $|B|$
is equal to $1043 \times 1043$.  There are only two possible
bracketings to choose from, so again $V$ is binary.  The training
corpus consists of about 35,000 two-word
compounds, giving $m = 35,000$ and $L \approx 1$ million.
 The accuracy reported is 75\%.

Resnik and Hearst (1993) aim to enhance Hindle and
Rooth's (1993) work by incorporating
information about the head noun of the prepositional
phrase in question.  Thus $B$ is now a product of
three spaces: the set of nouns and verbs, the set of
prepositions and the set of nouns.  To reduce the data
requirement, a freely available on-line thesaurus,
called WordNet is used (Beckwith {\it et al.}, 1991).
WordNet groups words into synsets, categories of synonymous
words.  These synsets are arranged in a taxonomy, so that
every word is also provided with a list of hypernyms.
The system then adds together the frequency counts for nouns
within a synset, providing more data about each.
This reduces the number of bins, since it is the synsets
which are taken as indicators rather than individual words.
If we assume roughly 1000 synsets,
1000 verbs and 100 prepositions, then $|B|
=  200$ million.  $V$ is still binary.  Their training corpus
is an ``order of magnitude smaller than'' Hindle and Rooth's,
so $m$ is around 100,000.  Unlike Hindle
and Rooth's, their corpus is parsed, which should give
better results.  Interestingly, they combine
evidence from large groups of synsets within WordNet's
hypernym hierarchy using a t-test.  This causes
the effective number of synsets for nouns to be reduced,
perhaps by as much as a factor of 10 (thus $|B| \approx 20$ million).
I will therefore assume that $L \approx 20$ million.  Even
given the additional information about the head
noun of the prepositional phrase, the accuracy reported
fails to improve on that of Hindle and Rooth,
being 78\%.  It is possible that insufficient training
data is the cause of this shortfall.

\begin{table*}
\begin{center}
\begin{tabular}{|c|c|c|c|c|c|c|}\hline
System & Training Source
& $m$ & $L$ & $m:L$ & Accuracy & Humans \\
\hline
\hline
Weischedel {\it et al.} & Manual Supervision
& 4M & 100k & 40 & 97\% &  ($\leq$ 97\%)
\protect\footnotemark \\
Yarowsky & Unsupervised
& 1G & 100M & 10 & 92\% & -\\
Hindle \& Rooth & Automatic Supervision
& 1.2M & $\sim$1M & $\sim$1 & 80\% & 85--88\%\\
Lauer & Automatic Supervision
& 35k & 1M & 0.035 & 75\% & ($\leq$ 80\%)
\protect\footnotemark \\
Resnik \& Hearst & Manual Supervision
& 100k & $\geq$20M & $\leq$0.005 & 78\% & ($\leq$ 92\%)
\protect\footnotemark \\
\hline
\end{tabular}
\end{center}
\caption{Summary of a sample of statistical NLP systems}
\label{summary_table}
\end{table*}
\addtocounter{footnote}{-2}
\footnotetext{Brackets indicate measured on different data
and/or under different conditions.}
\addtocounter{footnote}{1}
\footnotetext{Reported in Resnik(1993).}
\addtocounter{footnote}{1}
\footnotetext{Reported in Dras and Lauer(1993).}

Table \ref{summary_table} shows a summary of the above
systems, ordered on the ratio $m:L$.  A strong
correlation is evident between the value of this ratio
and the success rate.  This suggests that the success
of a statistically based system is strongly dependent
on the confidence permitted by the training set size
as measured by this ratio.

\subsection{An Important Trade-off}

The model formulated above and the empirical data presented
support a number of qualitative
inferences about the potential of systems given a fixed
training set size.  Because training data will
always be limited, such reasoning is an important part
of system design.  Therefore before turning to
some quantitative analyses, I will examine a few such
inferences.

The most important of these is in regard to linguistic
sophistication, that is the degree to which the
system uses knowledge of the patterns of language.
 This kind of knowledge is extremely important,
since it often allows just the right distinctions to
be made.  More simplistic systems will inevitably
assign one choice to two different inputs
because their linguistic knowledge fails to support a
distinction.  Therefore, it seems desirable to
incorporate as much linguistic sophistication as
possible.  While this is a tempting direction to take
for improving system performance, there is a
barrier.

Consider, for instance, the effect on data requirements
of incorporating new indicators.  Each indicator
increases the number of distinctions which the system
can make.  For example, Resnik and Hearst (1993)
take into account the object of the preposition. In
doing so, they distinguish cases which Hindle and Rooth (1993)
did not.  As a result, the number of cases their system
considers is substantially larger than those considered
by Hindle and Rooth's.  In terms of the framework, Resnik
and Hearst have many more bins than Hindle and Rooth.

It is easy to see that incorporating a new indicator
increases the number of bins combinatorially.
The size of $B$ is multiplied by the
range of the new indicator.  This results in the ratio $m:L$
falling by the same factor, which, as I have argued above, can be
detrimental to the overall accuracy.

The situation is worse still if the training set is not hand annotated.
In this case, introducing the new indicator creates additional ambiguity
in the training set since the value of the new indicator must be
determined for each training example.  This
effectively decreases the number of training instances
resulting in a further decrease in $m:L$.

Thus, linguistic sophistication presents a trade-off
between accuracy and data sparseness.  It is a
balance between poor modeling of the language and insufficient
data for accurate statistics.  If we are to
strike a satisfactory compromise, we need a strong
theory of data requirements and ways to make more
economic use of data.

One such method is termed {\em conceptual association},
as defined in Resnik and Hearst (1993).  By
collecting statistics based on concepts rather than
individual words, the number of bins is usually
reduced.  The idea is to generalise findings about
words to cover other words which have the same
meaning.  The advantages of this approach are extensively
argued in Resnik (1993) and the method is
used in Lauer (1994).  While concepts can help, the
ambiguity introduced (namely what concept does a
given word belong to) may undermine the increased accuracy.
  Further work is needed to establish the
effects on data requirements of employing this strategy.

A novel extension to this approach that has not yet
been employed, would be to collect statistics at
various levels of granularity.  Statistics computed on
counts of individual words would provide fine sensitivity,
while statistics computed on counts of a small set of
semantic primitives (such as ANIMATE, ABSTRACT, etc.)
would provide the coarsest evidence.  As many levels
as desired between these two extremes could be employed
in this way.  The level used to make
each choice could then be selected according
to the degree of confidence available at each level.
If insufficient data has been seen to allow a confident
selection at one level, a coarser grained level
would be tried.  Resnik and Hearst (1993) seem to be
simulating this when they perform a t-test across
all levels of the WordNet hierarchy.

\section{First Steps Towards a \mbox{Theory}} \label{theory}
\subsection{A Simple Learning Scheme}

In this section I shall establish some lower bounds
on the accuracy of a simple training scheme within
the framework developed.  The mathematics presented in
Sections \ref{ss_empty} through \ref{ss_nonempty}
was for the most part developed by Mark Johnson of Brown
University and completed by the author.  I wish to thank
him for his many communications in this regard.

Let $t(b,c) = \{ (b, v) | v \in V, (b, v) \in c \}$,
the training instances in a corpus $c$ that fall into bin $b$.

Let $f(v,t) = |\{ (b, v) | b \in B, (b, v) \in t \}|$,
the frequency of the value $v$ in the set of
training instances, $t$.

Let $\mbox{\it mode}(b,c) = \mbox{argmax}_{v \in V} f(v, t(b,c))$,
the most common value for instances from a corpus $c$ in a bin $b$.
Where several values have equal frequencies, one should be
chosen at random.

Define the learning algorithm such that:
\begin{displaymath}
P_{c}(b, v) = \left\{ \begin{array}{ll}
       1 & \mbox{if $v = \mbox{\it mode}(b,c)$} \\
	   0 & \mbox{otherwise} \end{array} \right.
\end{displaymath}
Since each bin has only one value with non-zero
probability, $V$ is effectively a binary set (either
the instance has the non-zero value or it does not).
Thus, $L = |B|$.  Notice also that the value assigned
highest probability by $P_{c}$ is the one most
frequently falling into the bin.  That is,
$v_{\mbox{\it mode}}(P_{c},b) = \mbox{\it mode}(b, c)$.

Two possible cases arise when the analyser is faced
with making a decision on the basis of some
indication, $b$.  Either the corpus contains no occurrences
of $(b, v)$ for any value $v \in V$ (Case A)
or there is some training data which falls into the
bin (Case B).

\subsection{Empty Bins} \label{ss_empty}

Case A arises when none of the training instances fall
into the bin.  Let $p_{b}$ denote \mbox{$\Pr(I=b)$}.  The
probability of bin $b$ being empty after training on a
randomly selected corpus is $(1-p_{b})^{m}$.  Thus the
probability over all test inputs of there being no training
data in the bin for that input is:
\begin{displaymath}
w = \sum_{b \in B} p_{b}(1-p_{b})^{m}
\end{displaymath}
Since we know that $\sum_{b \in B} p_{b} = 1$, it is
possible to show that the maximum for $w$
occurs when $\forall b \in B$ $p_{b} = \frac{1}{|B|}$.
 Therefore:
\begin{displaymath}
w \leq (1-\frac{1}{|B|})^{m} \leq e^{-m/{|B|}}
\end{displaymath}
So for even quite small values of $m/{|B|}$, the
probability that any given test sample falls
into a bin for which we received no training samples is
very low.  For example, when $m/{|B|}\geq 3$,
they occur in less than 5\% of inputs.

\subsection{Non-empty Bins} \label{ss_nonempty}

In Case B, we have at least one instance in the corpus
for the given bin.  Let \mbox{$n = |t(b, c)| \geq
1$}.  An optimal function, $p_{\mbox{\it opt}}$, will
be one which chooses for bin $b$ the value $v$ that maximises
$\Pr(J=v | I=b)$.
Let $v_{\mbox{\it opt}}(b) = \mbox{argmax}_{v \in V} \Pr(J=v |
I=b)$, the most likely value in bin $b$.
Let $q(b) = \Pr(J=v_{\mbox{\it opt}}(b) | I=b)$, the
probability of this value given an instance in bin $b$.
Notice from equation (\ref{eer_def}) that the expected
error rate is minimised when $\forall b \in B$
$v_{\mbox{\it mode}}(p,b) =  v_{\mbox{\it opt}}(b)$.  Therefore:
\begin{eqnarray*}
\lefteqn{r_{\mbox{\it opt}}} \\
& = & \sum_{b \in B} \Pr(I=b) (\!\!\!\!\sum_{v \in
V \setminus \{v_{\mbox{\it opt}}(b) \}}
\!\!\!\!\!\!\!\!\!\!\!\!\!\!\!\!\!\Pr(J=v | I=b) ) \\
& = & \sum_{b \in B} \Pr(I=b) (\Pr(J \neq v_{\mbox{\it opt}}(b)
| I=b) ) \\
& = & \sum_{b \in B} \Pr(I=b) (1-q(b)) \\
\end{eqnarray*}

Since $r_{\mbox{\it opt}}$ is the best possible error rate,
it follows that $q(b)$ must be high
for most bins if the system is to work at all.  Therefore,
$v_{\mbox{\it opt}}(b)$ should be a frequent value in each bin.
Now if more than half of the instances in a bin have the value
$v_{opt}$, then this must be the most common value in the bin.
Thus, if $f(v_{\mbox{\it opt}}(b),t(b,c)) > \frac{n}{2}$, then
$\mbox{\it mode}(b,c) = v_{\mbox{\it opt}}(b)$.
So by computing the probability of
$f(v_{\mbox{\it opt}}(b),t(b,c))$ $> \frac{n}{2}$, we can
obtain a lower bound for the accuracy on bin $b$.
\footnote{The argument shown holds for all odd $n$.
A variation of the argument that bounds the expected accuracy
for all even $n$ is simple to construct.}

\begin{eqnarray*}
\lefteqn{\Pr(v_{\mbox{\it opt}}(b) = v_{\mbox{\it mode}}(p,b) | I=b)}  \\
 & = & \Pr(\mbox{\it mode}(b,c) = v_{\mbox{\it opt}}(b) | I=b) \\
 & \geq & \sum_{i=\frac{n+1}{2}}^{n}
\Pr(f(v_{\mbox{\it opt}}(b), t(b,c))=i | I=b) \\
 & = & \sum_{i=\frac{n+1}{2}}^{n}
           {n \choose i} (1-q(b))^{n-i} q(b)^{i} \\
\end{eqnarray*}
\begin{eqnarray*}
\lefteqn{ \mbox{Thus} \!\!\!\!\!\!
\sum_{v \in V \setminus \{ v_{\mbox{\it mode}}(p,b) \} }
    \!\!\!\!\!\!\Pr(J=v | I=b) } \\
 & = & 1 - \Pr(J=v_{\mbox{\it mode}}(p,b) | I=b) \\
 & \leq & 1 - q(b) \sum_{i=\frac{n+1}{2}}^{n}
        {n \choose i} 
        (1-q(b))^{n-i} q(b)^{i} \\
\end{eqnarray*}
\begin{eqnarray*}
\lefteqn{ \mbox{ since } \Pr(J=v_{\mbox{\it mode}}(p,b) | I=b) } \\
& \geq & \Pr(J=v_{\mbox{\it opt}}(b) | I=b) \\
& & \hspace{0.3in} \Pr(v_{\mbox{\it opt}}(b) = v_{\mbox{\it mode}}(p,b) | I=b)
\end{eqnarray*}
\begin{eqnarray*}
\lefteqn{ \mbox{Let} U_{n}(b) = } \\
& & 1-q(b) \sum_{i=\frac{n+1}{2}}^{n}
                  {n \choose i} (1-q(b))^{n-i} q(b)^{i}
\end{eqnarray*}

This is an upper bound on the expected error rate for
bin $b$. \footnote{For odd $n$.  When $n$ is even it can
be shown that is $U_{n-1}(b)$ is an upper bound.}
So:
\begin{displaymath}
R(P_{c}) \leq \sum_{b \in B} \Pr(I=b) U_{n}(b)
\end{displaymath}
As noted above $r_{\mbox{\it opt}} =  \sum_{b \in B} \Pr(I=b)
(1-q(b))$.  A comparison between the upper bound
$U_{n}(b)$ and the optimal error rate $1-q(b)$ shows that
for reasonably high values of $q(b)$ that $U_{n}(b)$ is close to
$1-q(b)$.  For example, when $q(b) \geq 0.9$, $U_{3}(b)
\leq 1.26(1-q(b))$ and
$U_{5}(b) \leq 1.08(1-q(b))$.

In fact, we can derive a bound for any $n$ as follows:
\begin{eqnarray}
U_{n}(b) & \leq & U_{1}(b) \nonumber \\
 & = & 1 - q(b)q(b)  \label{u_one_eqn} \\
 & = & (1+ q(b))(1 - q(b)) \nonumber \\
 & \leq & 2 (1-q(b)) \nonumber \\
 & = & 2 \Pr(J \neq v_{\mbox{\it opt}}(b) | I=b) \nonumber
\end{eqnarray}

Thus in all bins which have training instances in the
corpus, the expected error rate for the bin never
exceeds twice the optimal error rate for that bin.
It is interesting to note that Hindle and Rooth's (1993)
system has roughly one instance per bin and an optimal
error rate of 12\% (assuming the human accuracy of 88\%
is optimal), so that equation (\ref{u_one_eqn})
predicts a lower bound of 77\% accuracy.  This is just
less than the 80\% observed.  \footnote{I wish to thank
Eugene Charniak for pointing out this fact.}

When 5 instances from the training corpus fall into the
bin, the expected error rate approaches the optimal error
rate closely and when there are an average of 3
instances per bin, very few bins do not have instances
from the training corpus.  So, in general it appears that
3--5 instances per bin will be sufficient.

\subsection{Skewed Bins}

An obvious question is why systems such as Lauer (1994)
and Resnik and Hearst (1993) work at all
given that far less than one instance is expected for
each slot.  One possible answer is that different bins
have widely differing frequencies.  The system quickly
learns about the most frequent cases at the
expense of less frequent ones.

This can be modeled by considering the different
distributions, $p_{b}$ defined for Case A above.
In that analysis, the probability of encountering an
empty bin was maximised over all possible distributions.
However, if something is known about the distribution,
in principle a tighter bound is possible.  For example,
suppose some fraction of the bins have very low
probability.  That is, $\exists B' \subseteq B$ such that
$\sum_{b \in B'} p_{b} = c$ for some small c.
Let $B'' = B \setminus B'$ and $\beta = \frac{|B''|}{|B|}$.
Now:
\begin{eqnarray*}
w & = & \sum_{b \in B} p_{b} (1-p_{b})^{m} \\
 & = & \sum_{b \in B'} p_{b} (1-p_{b})^{m} + \sum_{b
\in B''} p_{b} (1-p_{b})^{m} \\
 & \leq & \sum_{b \in B'} p_{b} + \sum_{b \in B''} p_{b}
(1-p_{b})^{m} \\
 & \leq & c + \sum_{b \in B''} p_{b} (1-p_{b})^{m} \\
\end{eqnarray*}
Now the second term is maximised when $\forall b \in B''$ $p_{b}
=  \frac{1-c}{|B''|} = \frac{1-c}{\beta |B|}$.
So letting $\beta_{c} = \frac{\beta}{1-c}$:
\begin{eqnarray*}
w & \leq & c + (1 - \frac{1-c}{\beta |B|})^{m} \\
  & = & c + (1 - \frac{1}{\beta_{c} |B|})^{m} \\
  & \leq & c + e^{-m/{\beta_{c} |B|}} \\
\end{eqnarray*}
So knowing a pair of values $c$ and $\beta$ is a useful,
if primitive, means of lowering the upper
bound on data requirements.  Since the distribution
of bins does not depend on the values we are
seeking to learn, it should be possible to develop
simple techniques for estimating values of $c$ and
$\beta$.

\subsection{Future Work}

A great deal of work remains to be done.  I will mention
only a few directions where the work begs
to be extended.  First, the mathematical model doesn't
capture several aspects of existing models, such as maximising
probabilities over sequences of words and combining evidence
from multiple sources.  Second, the simple learning algorithm
presented differs from those used in practice in several
ways.  It would be useful to explore the
relationship between the algorithm I have proposed
and others in existing statistical methods (for
example, the backing off method in Katz, 1987).  Third,
smoothing is frequently used to alleviate data
sparseness (see Dagan {\it et al.}, 1993), but the model does
not include any means to represent the process
of smoothing.  Finally, almost all statistical NLP
systems deal with some noise in the training data.
This is especially important in systems like Yarowsky
(1992) where training is unsupervised.  The
mathematical results need to be extended to reflect
noisy training data and to support reasoning about
the sensitivity of data requirements to noise.

\section{Conclusion}

In this paper I have indicated the lack of a general
theory of data requirements within the field of
statistical NLP.  As a first step in the development
of such a theory I have presented a framework
for statistical NLP systems.  I have shown how several
prominent works in the field fit this model and demonstrated
a number of mathematical results which
support inferences about data requirements.  I believe
this represents a significant first step along the
road to a better understanding of when and how statistical
NLP methods can be applied.

\section{Acknowledgments}

Without Mark Johnson's interest and collaboration, this paper
would not exist.  Many of the key ideas originate with him
and I am indebted to him for his patience and attention.
The work has also benefited from assistance from Richard Buckland,
Robert Dale, Mark Dras, Mike Johnson and Steven Sommer.
Financial support is gratefully acknowledged from the Australian
Government and the Microsoft Institute.

\end{document}